\documentclass{article}

\usepackage{arxiv}
\usepackage{tikz}
\usepackage{bm}
\usepackage{amssymb}
\usepackage{amsmath}
\usepackage[utf8]{inputenc} 
\usepackage[T1]{fontenc}    
\usepackage{hyperref}       
\usepackage{url}            
\usepackage{booktabs}       
\usepackage{amsfonts}       
\usepackage{nicefrac}       
\usepackage{microtype}      
\usepackage{lipsum}		
\usepackage{graphicx}
\usepackage{doi}

\newcommand{\ie}{\textit{i}.\textit{e}.}
\newcommand{\eg}{\textit{e}.\textit{g}.}

\title{Generative Binary Memory: \\
Pseudo-Replay Class-Incremental Learning on Binarized Embeddings}

\author{ \href{https://orcid.org/0009-0000-5109-6602}{\includegraphics[scale=0.06]{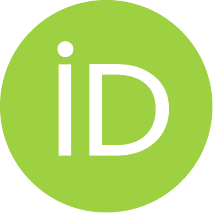}\hspace{1mm}Yanis BASSO-BERT}\\
	Univ. Grenoble Alpes, CEA, List\\
	F-38000 Grenoble\\
	France \\
	\texttt{yanis.bassobert@cea.fr} \\
	\And
	\href{https://orcid.org/0000-0003-2400-165X}{\includegraphics[scale=0.06]{orcid.pdf}\hspace{1mm}Anca MOLNOS}\\
	Univ. Grenoble Alpes, CEA, List\\
	F-38000 Grenoble\\
	France \\
	\texttt{anca.molnos@cea.fr} \\
	\And
	\href{https://orcid.org/0000-0002-7260-2786}{\includegraphics[scale=0.06]{orcid.pdf}\hspace{1mm}Romain LEMAIRE}\\
	Univ. Grenoble Alpes, CEA, List\\
	F-38000 Grenoble\\
	France \\
	\texttt{romain.lemaire@cea.fr} \\
	\And
	\href{https://orcid.org/0000-0001-8925-0441}{\includegraphics[scale=0.06]{orcid.pdf}\hspace{1mm}William GUICQUERO}\\
	Univ. Grenoble Alpes, CEA, Leti\\
	F-38000 Grenoble\\
	France \\
	\texttt{william.guicquero@cea.fr} \\
	\And
	\href{https://orcid.org/0000-0002-0145-2186}{\includegraphics[scale=0.06]{orcid.pdf}\hspace{1mm}Antoine DUPRET}\\
	Univ. Grenoble Alpes, CEA, Leti\\
	F-38000 Grenoble\\
	France \\
	\texttt{antoine.dupret@cea.fr} \\
}

\renewcommand{\shorttitle}{Generative Binary Memory}

\hypersetup{
pdftitle={Towards Experience Replay for Class-Incremental Learning in Fully-Binary Networks},
pdfsubject={cs.ai},
pdfauthor={Yanis BASSO-BERT, Anca MOLNOS, Romain LEMAIRE, William GUICQUERO, Antoine DUPRET},
pdfkeywords={Continual Learning, Class-Incremental Learning, Replay Methods, Binary Neural Network, Quantization},
}

\begin{document}
\maketitle

\begin{abstract}
In dynamic environments where new concepts continuously emerge, Deep Neural Networks (DNNs) must adapt by learning new classes while retaining previously acquired ones. This challenge is addressed by Class-Incremental Learning (CIL). This paper introduces Generative Binary Memory (GBM), a novel CIL pseudo-replay approach which generates synthetic binary pseudo-exemplars. Relying on Bernoulli Mixture Models (BMMs), GBM effectively models the multi-modal characteristics of class distributions, in a latent, binary space. With a specifically-designed feature binarizer, our approach applies to any conventional DNN. GBM also natively supports Binary Neural Networks (BNNs) for highly-constrained model sizes in embedded systems. The experimental results demonstrate that GBM achieves higher than state-of-the-art average accuracy on CIFAR100 ($+2.9\%$) and TinyImageNet ($+1.5\%$) for a ResNet-18 equipped with our binarizer. GBM also outperforms emerging CIL methods for BNNs, with $+3.1\%$ in final accuracy and $\times 4.7$ memory reduction, on CORE50.
\end{abstract}


\section{Introduction}
\label{sec:intro}

\begin{figure*}[t]
  \centering
   \includegraphics[trim=0 436 0 0,clip, scale = 0.97]{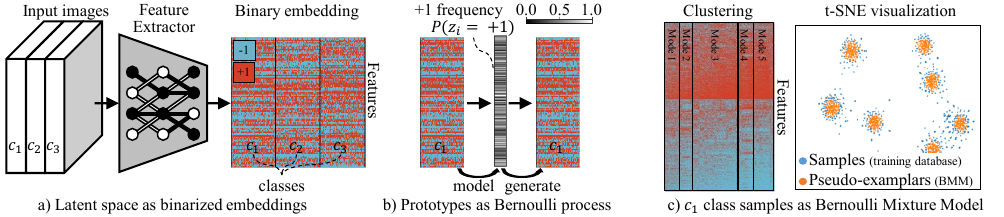}
   \caption{Motivations of our pseudo-replay CIL approach based on Bernoulli Mixture Model (BMM). a) Illustration of per-class features correlation on binary embedding. b) Illustration of a single prototype encoding and pseudo-examplars generation. Prototype features is computed as the +1's frequency in a class embedding. Pseudo-examplars are generated from the prototype as a Bernoulli process. c) Qualitative investigation of the binary distribution: (left) apparition of modes when reorganising samples and features with hierarchical clustering, (right) t-SNE visualization~\cite{van2008visualizing} of training samples and pseudo-exemplars generated from a BMM.}  
   \label{fig:1_motivation}
\end{figure*}

Many real-world applications face changing contexts, with continually emerging new concepts and categories \cite{hadsell2020embracing}. A Deep Neural Network (DNN) deployed in such dynamic environments should thus possess the ability to learn new classes, while preserving prior knowledge of past class, a learning paradigm referred as Class-Incremental Learning (CIL) \cite{kirkpatrick2017overcoming}. However, without any precautions, DNNs tend to forget previously learned patterns when being fine-tuned on new data, a phenomenon known as catastrophic forgetting~\cite{mccloskey1989catastrophic}. Numerous methods now exist~\cite{wang2024comprehensive} to effectively learn from a new set of classes, also denoted as a new task, without forgetting past classes.

Among the available methods, exemplars-based approaches have shown superior performance by storing and replaying a small portion of past examples, i.e., exemplars, during training \cite{wang2024comprehensive}. However, this advantage comes at the cost of increased memory requirements. Storing even a small subset of prior examples demands additional memory resources. This poses a significant challenge to CIL deployment on embedded systems, where memory is a scarce resource~\cite{ravaglia2021tinyml}. To address this, considerable efforts have focused on reducing memory size, primarily by compressing the stored exemplars \cite{wang2021memory} or storing exemplars in an embedding space, method also known as latent replay (LR)~\cite{pellegrini2020latent}.

More recently, LR is extended to Binary Neural Networks (BNNs)~\cite{hubara2016binarized}.
BNNs are a valuable approach, as they significantly reduce memory demands, computational complexity, and energy consumption through binary weights and activations~\cite{yuan2023comprehensive}. Furthermore, LR on BNNs has only a small memory overhead because of the binary nature of the embedding~\cite{Vorabbi24,basso2024class}.

This work takes a step further by generating synthetic, binary, pseudo-exemplars on-the-fly, rather than replaying stored binary exemplars. A generative model offers the potential to significantly reduce memory footprint and to avoid overfitting by interpolating between past training data. Existing exemplar-free CIL methods (\eg, EFCIL~\cite{petit2023fetril}) compute and store only a representative prototype for each past class. Synthetic pseudo-examplars are generated around prototypes by data augmentation heuristics. Building upon EFCIL, we investigate the utilisation of binary embeddings, as briefly presented in Figure~\ref{fig:1_motivation}.

As illustrated in Figure~\ref{fig:1_motivation}a), in a VGG-like BNN we observe that binary features (with values in $\{-1,+1\}$) tend to cluster by class, with some features having higher flip rates. The $+1$'s frequency for each feature may constitute a representative prototype of a class' binary embedding distribution. The class distributions can then be synthesized by Bernoulli processes. Moreover, we observed that multiple modes appear when applying hierarchical clustering~\cite{nielsen2016introduction} to class embedding (Figure~\ref{fig:1_motivation}c)), suggesting that a single prototype may not capture the full diversity of a class. To address this, the Bernoulli Mixture Model (BMM) offers a statistical generative framework to model binary vectors with multi-modal characteristics (Figure~\ref{fig:1_motivation}c)). This paper brings the following contributions: 
\begin{itemize}
    \item Generative Binary Memory (GBM), a pseudo-replay CIL method on binarized embeddings, based on BMMs. 
    \item Dedicated embedding binarizers making our GBM compatible with any features extractor and DNN. 
    \item Experimental results on a ResNet-18~\cite{he2016deep} demonstrating GBM improves state-of-the-art performance in comparison with other prototypes methods, \eg, by $+1.5\%$ against FeTril~\cite{petit2023fetril} on the challenging dataset of TinyImageNet with 20 incremental tasks.
    \item Experimental results on GBM with BNNs, managing CIL under a limited memory budget dedicated to tiny DNN models, outperforming emerging work \cite{basso2024class, Vorabbi24} with $+3.1\%$ in final accuracy and $\times 4.7$ memory reduction, on CORE50 benchmark.
\end{itemize}\newcommand{\yanis}[1]{\textcolor{black}{#1}}

\section{Related Work}
\label{sec:related_works}

\textbf{Continual Learning.} Among the continual learning problems in the literature~\cite{wang2024comprehensive}, our focus is on Class Incremental Learning, CIL~\cite{rebuffi2017icarl}, whose goal is to learn a unified classifier without relying on task labels. CIL inherently involves a plasticity-stability trade-off, where the model must remain flexible to learn new tasks and stable enough to preserve previously acquired knowledge. Various methods exist, including architectural-based approaches~\cite{yan2021dynamically}, regularization-based techniques like Elastic Weight Consolidation (EWC)~\cite{kirkpatrick2017overcoming} and Learning without Forgetting (LwF)~\cite{li2017learning}, and replay-based strategies~\cite{rebuffi2017icarl}. Replay methods, such as iCaRL~\cite{rebuffi2017icarl}, EEIL~\cite{castro2018end}, and LUCIR~\cite{hou2019learning}, are effective but introduce the additional memory overhead related to past tasks exemplars. LR methods~\cite{pellegrini2020latent} address the memory issues, storing latent features instead of raw data.

Furthermore, prototype-based methods rely on class centroids and distributions modeling to generate synthetic data. For instance, PASS~\cite{zhu2021prototype} uses isotropic augmentation based on the trace of the covariance matrix, while IL2A~\cite{zhu2021class} accounts for feature inter-dependencies estimations using a multivariate Gaussian approximation. More recently, FeTril~\cite{petit2023fetril} augments the distribution of old classes based on new classes statistics.
Knowledge distillation~\cite{gou2021knowledge} can also cap drift from previously learned distributions ~\cite{zhu2021class, zhu2021prototype}.
 
Like FeTril, we address a system with a fixed feature extractor to ensure a stable representation, while retraining only the classifier. Our proposed method further differentiates from other prototype-based approaches by (1) working on categorical binary features rather than continuous (2) handling the multi-modality of class distributions using a mixture model, and (3) applying prototype quantization.

\textbf{Bernoulli Mixture Model in CIL.} BMMs have been widely studied to model multidimensional categorical data~\cite{bishop2006pattern,juan2004initialisation,li2016conditional,lazarsfeld1968latent}, with applications in clustering~\cite{najafi2019reliableclusteringbernoullimixture}, classification~\cite{alalyan2020hybrid,gimenez2014discriminative}, and dimensionality reduction~\cite{saeed2013machine}. More recently, BMMs are combined with deep neural networks, particularly for parameterizing priors in variational auto-encoders~\cite{wu2019deep,loaiza2019continuous}, image retrieval~\cite{amato2016combining,amato2018aggregating} and multi-instance learning~\cite{yang2020multiple}.  
BMMs have not been yet applied to CIL, unlike Gaussian Mixture Models (GMMs)~\cite{bishop2006pattern}. 
For example, Gaussian Mixture Replay~\cite{pfulb2021overcoming} models and generates input images using GMMs, however it limits to simple datasets such as MNIST~\cite{lecun1998gradient}. 
Furthermore, MIX~\cite{korycki2024class} learns GMMs and uses them as classifiers, but performs well only when the model is pre-trained on the whole dataset.

\begin{figure*}[h!]
\centering
\includegraphics[trim=0 256 0 0,clip, scale = 0.8]{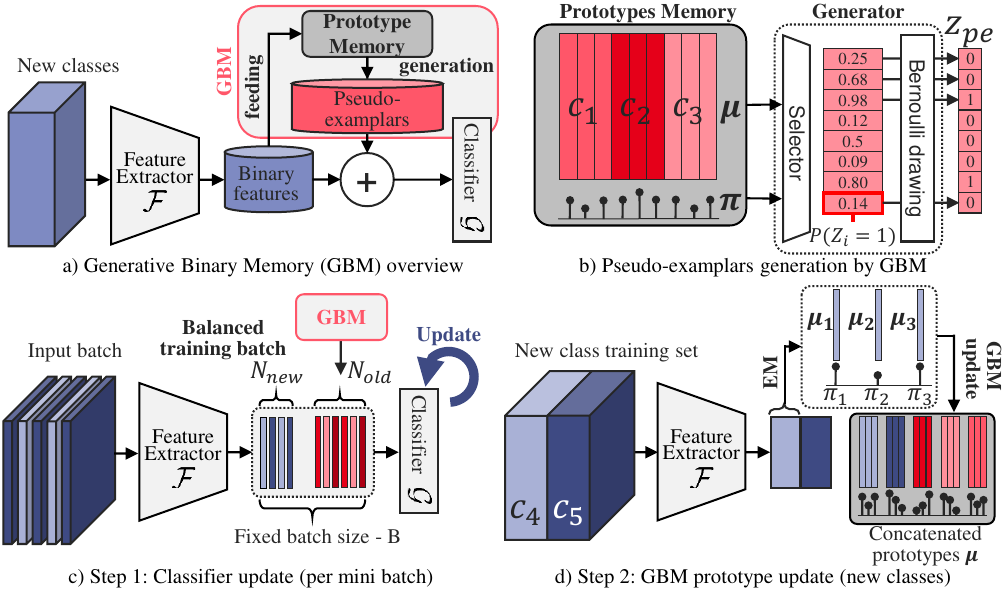}
\caption{Our proposed GBM method for Pseudo Replay on Binary embedding: a) overview
of our framework, b) generative procedure from the prototype memory, c) Classifier update on class-balanced batch between new class samples and past class pseudo-exemplars, and d) GBM update with per-class BMM modeling with EM algorithm and prototype concatenation is the memory.}
\label{fig:4_GBM_overview}
\end{figure*}

\textbf{Binary Neural Network in CIL.} BNNs have binary weights and activations (\eg,  $\{-1,+1\}$~\cite{yuan2023comprehensive}) and thus all scalar products are replaced with more efficient XNOR logic gates and bit-count operations~\cite{hubara2016binarized}. As a result, BNNs drastically reduce memory footprint, and energy consumption, making them highly efficient in terms of hardware implementation. Despite their non-differentiable nature, BNNs training through back-propagation is made possible by the Straight-Through Estimator (STE)~\cite{bengio2013estimating} in a Quantization-Aware Training (QAT) approach~\cite{kulkarni2022survey}. 
Nevertheless, BNNs' performance does not yet match the one of the corresponding conventional, real-valued networks. A way to narrow this gap is to keep some layers in floating-point precision~\cite{bannink2021larq,liu2018bi,liu2020reactnet}, leading to \textbf{Hybrid BNNs}. Another branch of research focuses on optimizing BNNs that entirely rely on binary arithmetic~\cite{ma2024b,basso2024class}, \ie \textbf{Fully BNNs}, to enable disruptive hardware implementations. 

Recent work also explores CIL for BNNs. Synaptic metaplasticity~\cite{laborieux2021synaptic} introduces a regularization function that computes weights' importance based on weights' magnitude during QAT, however is limited to simple datasets. LR on binary embedding is proposed in~\cite{Vorabbi24}, taking advantage of BNNs' compression. Finally, \cite{basso2024class} explores replay mechanisms in fully binary networks, showing the suitability of both experience and LR for ultra-low power devices. This work goes beyond replay methods by generating binary pseudo-exemplars based on intra-class correlations, to increase performance and further reduce CIL memory needs.

\section{Generative Binary Memory}
\label{sec:generative_binary_memory}

\textbf{CIL problem formulation.} The CIL problem involves learning a sequence of classification tasks, starting from an initial task, with training dataset $\mathcal{D}_0$, followed by $T$ incremental tasks, with training datasets $\mathcal{D}_1$,...,$\mathcal{D}_T$. $\mathcal{D}_t = \left\{(\bm{x_{t,i}},y_{t,i})\right\}_{i=1}^{N_t}$ is the training dataset that the model receives at task $t$. 
$\mathcal{D}_t$ consists in $N_t$ labeled samples, $\bm{x_{t,i}}$, $y_{t,i} \in \mathcal{C}_t$ denotes the class labels, and $\mathcal{C}_t$ is the set of classes for task $t$. There is no intersection between the classes of different tasks, meaning $\mathcal{C}_{i} \cap \mathcal{C}_{j} = \emptyset$ for $i \neq j$. Typically, at task $t$, the model is updated only on the training examples from the current task, $\mathcal{D}_t$ and evaluated on the classification problem containing all seen classes $\bigcup_{i=0}^t \mathcal{C}_i$.

\textbf{System overview.} We consider a neural network composed of a feature extractor $\mathcal{F}$ and a classifier $\mathcal{G}$. This neural network is extended with our Generative Binary Memory (GBM), as depicted in Figure~\ref{fig:4_GBM_overview}a). The GBM includes a prototype memory that: (1) can be updated when new classes appears, and (2) can generate synthetic pseudo-examplars (Figure~\ref{fig:4_GBM_overview}b)). $\mathcal{F}$ is trained during the initial task and then frozen. For each incremental task, the system is updated in two steps. First, $\mathcal{G}$ is re-trained on new training samples mixed with generated pseudo-exemplars associated to old classes (Figure~\ref{fig:4_GBM_overview}c)). Second, the GBM is updated with prototypes computed on the new classes training set. Prototypes and their mixing coefficients are computed per-class as BMM with an EM algorithm (Figure~\ref{fig:4_GBM_overview}d)).

In what follows, Section~\ref{sec:section-3.1} details how prototypes are computed, given the binary features. Section~\ref{sec:section-3.2} then describes the prototype memory update and the process of data generation from prototypes. Finally, while our method is motivated by CIL on BNNs, we extend its applicability to any neural network; for this two embedding binarizer blocks are proposed in Section~\ref{sec:section-3.3}. 

\subsection{Multi-prototype encoding with BMM}\label{sec:section-3.1}

At a given task $t$, GBM computes a multi-prototype representation of the $N$ training samples, $\bm{X} = \left\{\bm{x_{t,i}} \mid y_{t,i} = c\right\}$ of a given class $c \in C_t$. Let $\bm{Z} = \mathcal{F}(\bm{X})$ be the $N$ latent binary embedding set with dimension $D$, \ie   $\bm{Z}=\left\{\bm{z_i}\right\}_{i=1}^N \in \{0,1\}^{N\times D}$. Under the assumption that $\bm{Z}$ is generated by $K \in \mathbb{N}$ underlying Bernoulli processes, we model the distribution with a BMM parameterized by $\Phi = (\bm{\mu},\bm{\pi})$, where $\bm{\mu} = \{\bm{\mu_k}\}_{k=1}^K \in [0, 1]^{K\times D}$ denotes the $K$ Bernoulli probability vectors, namely prototypes, and $\bm{\pi} = \{\pi_k\}_{k=1}^K \in [0, 1]^{K \times 1}$ denotes the corresponding mixing coefficients, subject to $\sum_{k=1}^K \pi_k = 1$.

An EM algorithm~\cite{bishop2006pattern} estimates the parameters $\Phi$ from $\bm{Z}$. The EM algorithm iteratively refines $\Phi$ through two main steps: (1) an expectation step (E-step), and (2) a maximization step (M-step). In the E-step (Equation~\ref{eq:E-step}), the responsibilities $\gamma_{i,k}$, \ie, the probability that a binary embedding $\bm{z_i}$ was generated by component $k$, are defined as:

\begin{equation}
    \label{eq:E-step}
    \gamma_{i,k} = \frac{\pi_k \prod_{j=1}^D \mu_{k,j}^{z_{i,j}} \left( 1 - \mu_{k,j} \right)^{1 - z_{i,j}}}{\sum_{l=1}^K \pi_l \prod_{j=1}^D \mu_{l,j}^{z_{i,j}} \left( 1 - \mu_{l,j} \right)^{1 - z_{i,j}}}
\end{equation}

In the M-step (Equation \ref{eq:M-step}), the model parameters $\bm{\mu}$ and $\bm{\pi}$ are updated based on the computed responsibilities:

\begin{align}
\label{eq:M-step}
\pi_k = \frac{1}{N} \sum_{i=1}^N \gamma_{i,k} && \mu_{k,j} = \frac{\sum_{i=1}^N \gamma_{i,k} z_{i,j}}{\sum_{i=1}^N \gamma_{i,k}}
\end{align}

The BMM log-likelihood at iteration $s$ of the EM-algo $\ell(\Phi^{(s)})$ is further defined in Equation~\ref{eq:mstep} as:
\begin{equation}\label{eq:mstep}
    \ell(\Phi^{(s)}) = \sum_{i=1}^{N}\sum_{k=1}^K \gamma_{i,k} \log \left( \pi_k \prod_{j=1}^D \mu_{k,j}^{z_{i,j}} \left(1 - \mu_{k,j} \right)^{1 - z_{i,j}}\right)
\end{equation}

The algorithm stops when the relative change $\lvert \ell(\Phi^{(s)}) - \ell(\Phi^{(s-1)}) \rvert / \lvert \ell(\Phi^{(s)}) \rvert$, falls below a predefined threshold $\epsilon$.

\textbf{Initialization.} Initialization is critical for BMM to effectively converge and prevent ``pathological cases''~\cite{juan2004initialisation} such as prototype degeneration, \ie, when a prototype overfits on one single training point. We therefore initialize each prototype parameter $\bm{\mu}$ by calculating the centroid of $\bm{Z}$ and then perturbing each $\mu_k$ around this centroid with a per-feature standard deviation. The mixing coefficients $\bm{\pi}$ are initialized and fixed at $\pi_k = \frac{1}{K}$, and not trainable, unlike in Equation~\ref{eq:mstep}. This prevents prototype degeneration by encouraging a balanced prototype attribution during the E-step. Section~\ref{sec:evaluation_resnet} includes an experimental investigation of these BMM initialization choices. To further improve the initialization, we perform $N_{init}$ warm-up rounds with different initializations, running the EM algorithm for $N_{iter}$ iterations each. The initialization that yields the highest log-likelihood is selected, and then fully optimized up to the stopping criterion.

\textbf{Post-EM prototypes quantization.} While prototype methods are often described as memory-free~\cite{zhu2021prototype}, storing prototypes still requires memory, as highlighted in \cite{petit2023fetril}. To be more efficient than conventional methods that store latent binary samples, GBM prototypes can be uniformly quantized (on $q$ bits). 
Section~\ref{sec:FBNN} explores this, down to a prototype memory footprint close to the one of a few binary exemplars.

\subsection{GBM update and pseudo-exemplars generation}\label{sec:section-3.2}

A BMM, $\Phi_{new}$=$(\bm{\mu_{new}}, \bm{\pi_{new}})$, is computed for each new encountered class. The GBM is updated by appending the newly computed prototypes $\bm{\mu_{new}}$ to the existing set of prototypes $\bm{\mu}$, and the mixing coefficients $\bm{\pi_{new}}$ to $\bm{\pi}$. To ensure that the concatenated $\bm{\pi}$ accurately represents the drawing probability across all seen classes, $\bm{\pi}$ is normalized based on the effective number of training samples per class. 

During the classifier update on a new class, GBM generates pseudo-exemplars at each training batch. A pseudo-exemplar $(\bm{z_{pe}}, \bm{y_{pe}})$ is generated by first selecting a Bernoulli process $\bm{\mu_i}$ in $\bm{\mu}$ with selection probability $\bm{\pi}$. 
Then a realization is drawn from the Bernoulli process parameterized by $\bm{\mu_i}$ (Figure~\ref{fig:4_GBM_overview}b)).
Within the fixed training batch size $B$, the number of new training data $N_{new}$ and pseudo-exemplars $N_{old}$ is set proportionally to the number of new $n_{new}$ and previous classes $n_{old}$. This ensures an equal representation of all seen classes in batches (Equation \ref{eq:N}), to circumvent a possible imbalance dataset (Figure~\ref{fig:4_GBM_overview}c)).

\begin{align}
\label{eq:N}
N_{new} = \frac{B \times n_{new}}{n_{old} + n_{new}} && N_{old} = \frac{B \times n_{old}}{n_{old} + n_{new}}
\end{align}

\subsection{Feature embedding binarization}\label{sec:section-3.3}

\begin{figure}[htbp]
\centering
\includegraphics[trim=0 180 0 0,clip, scale = 0.5]{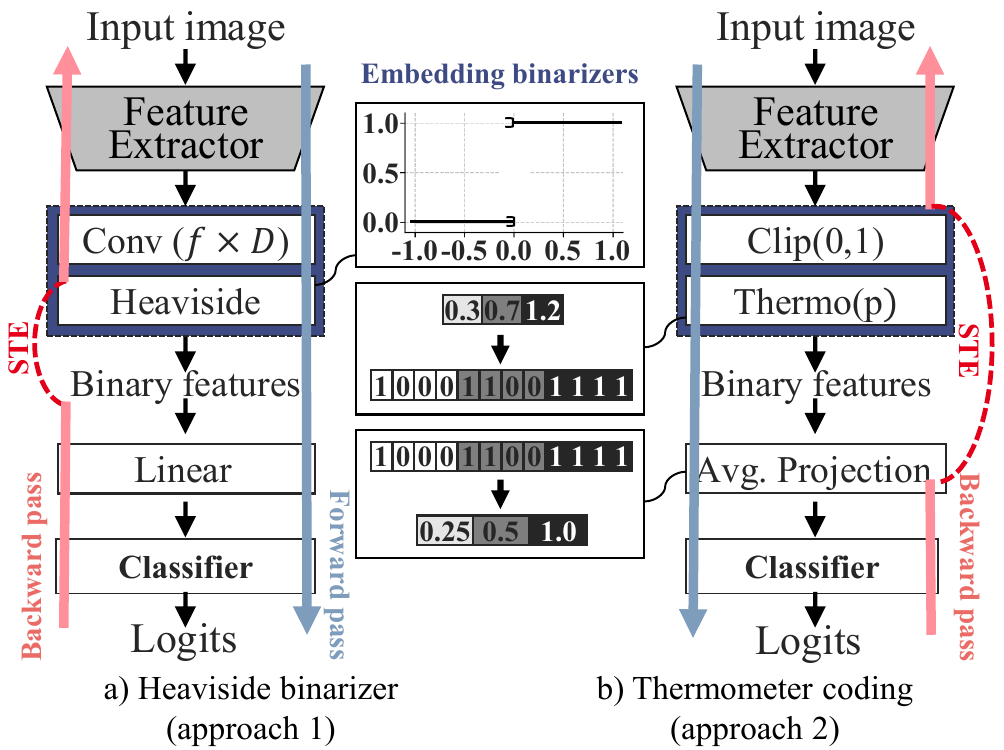}
\caption{The 2 proposed approaches for embedding binarization.}
\label{fig:4_embedding_binarization}
\end{figure}

GBM can be included in non-binary, conventional neural networks by adding an embedding binarizer and slightly adapting the training protocol of the initial task (Figure~\ref{fig:4_embedding_binarization}). Two approaches are possible: (1) enforcing $\mathcal{F}$ to output directly a binary vector on its last layer; (2) coding learned full precision features into a binary vector. 

\textbf{Heaviside binarizer.} For the first approach we propose to extend $\mathcal{F}$ with a point-wise convolution with $f\times D$ filters, followed by an Heaviside step activation for binary quantization. We train the model using the Straight-Through Estimator (STE)~\cite{bengio2013estimating} to compute the gradients in the Heaviside's backward computation, to learn a binary feature representation. The binary embedding size is controlled by a factor $f$, as investigated in Section~\ref{sec:evaluation_resnet}. 

\textbf{Thermometer binarizer.} For the second approach, a thermometer representation~\cite{buckman2018thermometer} encodes the real-valued $D$-dimensional embedding. A thermometer code represents an integer number on $p$ bits, where the number of consecutive ones corresponds to the encoded value. Assuming that features values are within $[0,1]$, we can uniformly quantize the interval $[0,1]$ on $p$ values and concatenate the associated thermocodes in one unique $p\times D$-dimensional binary vector. Equation ~\ref{eq:thermocoding} describes the thermode transfomation $\hat{\bm{z}}$ of a embedding vector $\bm{z}$. $z_i$ refers to the $i$-th feature.

\begin{equation}\label{eq:thermocoding}
    \hat{z}_{p\times i+j} = 
        \begin{cases}
                1 & \text{if } j \times \lfloor clip(z_{i}) \times p \rfloor \geq 1\\
                0 & \text{otherwise}
        \end{cases}
\end{equation}

Features need to be converted back in the real-valued domain, before being fed to the classifier.
This conversion is done by averaging each thermocode segment (Equation ~\ref{eq:th_inverse}). 
\begin{equation}\label{eq:th_inverse}
    z_{i} = \frac{1}{p}\sum_{j=0}^{p-1}  \hat{z}_{p \times i+j}
\end{equation}

 Since feature values may not be between 0 and 1, the output of $\mathcal{F}$ needs to be clipped in $[0, 1]$. However, to avoid too much activation saturation, a second training phase is performed with an STE from $\mathcal{F}$'s outputs to $\mathcal{G}$'s inputs. This way, $\mathcal{F}$ can learn to adjust its output dynamic range to the saturation and thermometer coding.

\section{Evaluation on ResNet architecture}
\label{sec:evaluation_resnet}

In this section, we evaluate the approach on a conventional ResNet, enhanced by our proposed embedding binarizers. 
This allows to compare with state-of-the-art methods in CIL, assess the generalizability of our technique, and test its robustness across various hyperparameters. Specifically, ResNet-18~\cite{he2016deep} is used as $\mathcal{F}$, which enables us to evaluate the applicability of GBM for pseudo-replay on rich features. 
The $\mathcal{F}$ terminates at the output of the average-pooling layer with an output dimension of $D$=512. The classifier $\mathcal{G}$ consists of a linear projection.

\subsection{Evaluation set-up}

\textbf{Datasets and incremental setting.} GBM is evaluated on two common datasets in CIL: CIFAR100~\cite{krizhevsky2009learning} with 100 classes and TinyImageNet~\cite{le2015tiny} with 200 classes, using a fixed random order for classes. Typically, the initial task dataset $\mathcal{D}_0$ consists of half of the classes and the other half of the classes are divided across the incremental tasks. We study CIL in three settings, with $T$=5, 10, and 20 classes. For $T$=20 on CIFAR100, the initial task includes 40 classes instead of the usual 50.

\textbf{Compared Methods.} We compare GBM with the two proposed binarizers, \textit{$p$-bits thermometer}, $GBM^T_p$, and \textit{$f$-heaviside}, $GBM^H_f$, to EFCIL methods: LwF-MC~\cite{li2017learning}, EWC~\cite{kirkpatrick2017overcoming} which are memory-free and PASS~\cite{zhu2021prototype}, IL2A~\cite{zhu2021class}, FeTRIL~\cite{petit2023fetril} which store class-prototypes. We also compare to state-of-the-art exemplars-based method iCaRL-CNN~\cite{rebuffi2017icarl}, EEIL~\cite{castro2018end} and LUCIR~\cite{hou2019learning}. 

\textbf{Metrics.} We report the average accuracy. Average accuracy is the average of the $T+1$ task accuracy (test set) on all the classes that have already been learned so far. We also consider final train and test accuracy to better understand the influence of embedding binarizer blocks.

\textbf{Training and hyperparameters.} Following the training protocol in \cite{petit2023fetril}, we train ResNet-18 from scratch using a batch size of 128 and the SGD optimizer with a momentum of 0.9. The initial learning rate is 0.1, exponentially decaying by a factor of 0.1 every 50 epochs, for a total of 160 epochs. Data augmentation includes random horizontal flips, 4-pixel translations, and random contrast adjustments. For BMM computation, we use $N_{init}$=5 and $N_{iter}$=3 for warm-up, with $\epsilon$=$10^{-3}$ and a maximum of $n_{max}$=10 steps for stopping criteria. Results are averaged over 3 runs.
\newline

\subsection{Results and discussion}

\begin{table}[htbp]
\centering
\begin{tabular}{cllccclccc}
\hline
Category        & Method &  & \multicolumn{3}{c}{CIFAR-100}                          &  & \multicolumn{3}{c}{TinyImageNet}                    \\ \cline{4-6} \cline{8-10} 
               &    \ \ \ \ \ \ \ \ \ $T$=  & & 5   & 10  & 20  &  & 5   & 10  & 20  \\ \hline
\small Replay         & iCaRL               &  & 51.1 & 48.7 & 44.4 &  & 34.6 & 31.2 & 27.9 \\
$E$=20               & EEIL                    &  & 60.4 & 56.1 & 52.3 &  & 47.1 & 45.0 & 40.5 \\
               & LUCIR                   &  & 63.8 & 62.4 & 59.1 &  & 49.1 & 48.5 & 42.8 \\ \hline
\small Regularization & LwF-MC                  &  & 45.9  & 27.4  & 20.1  &  & 29.1  & 23.1  & 17.4  \\
$K$=0            & EWC                     &  & 24.5  & 21.2 & 15.9 &  & 18.8 & 15.8 & 12.4 \\ \hline
\small Prototype      & PASS                    &  & 63.5  & 61.8  & 58.1  &  & 49.6  & 47.3  & 42.1  \\
$K$=1            & IL2A                    &  & 66.0  & 60.3  & 57.9  &  & 47.3  & 44.7  & 40.0  \\
                & FeTril                      &  & \underline{66.3} & \underline{65.2} & 61.5             &  & \textbf{54.8}    & 53.1          & 52.2             \\
               & \textbf{$GBM^H_{2}$} &  & 64.5  & 64.3  & 61.1  &  & 53.4  & 52.7  & 52.2  \\
               & \textbf{$GBM^T_1$}      &  & 66.0  & 65.8  & 64.3  &  & 53.9  & 53.2  & 52.8  \\ \hline
\small Multiprototype & \textbf{$GBM^H_{2}$}     &  & 65.2             & 65.0             & \underline{61.7} &  & 54.0             & \underline{53.3}          & \underline{52.9} \\
$K$=8             & \textbf{$GBM^T_1$}          &  & \textbf{67.1}    & \textbf{67.0}    & \textbf{64.4}    &  & \underline{54.6} & \textbf{54.0} & \textbf{53.7}    \\ \hline
\end{tabular}
\caption{GBM average incremental accuracy compared to replay methods ($E$=20 exemplars per class), memory-free regularization methods and prototype methods. Best $GBM^T$ and $GBM^H$ configurations from Table~\ref{tab:investigation_binarization} are reported. \textbf{Best results}, \underline{second best results}.}
\label{tab:resnet_benchmark}
\end{table}

\textbf{Comparison with the State-of-the-art.} Table~\ref{tab:resnet_benchmark} reports the experimental results on CIFAR100 and TinyImageNet for the three incremental settings. The results indicate that our 8-prototypes $GBM^T_1$ has consistently better performance than state-of-the-art, on nearly every configuration.
Compared to the second-best method, FeTril, GBM's accuracy is $+1.5\%$ higher, on the most challenging case of TinyImageNet with $T$=20. Moreover our method maintains high final performance when the number of tasks increases, with a relative decrease of $1.6\%$ on TinyImageNet between $T$=5 and $T$=20, compared to a relative $4.7\%$ for FeTril. 

\begin{figure*}[htbp]
\centering
\includegraphics[trim=0 375 0 0,clip, scale = 0.97]{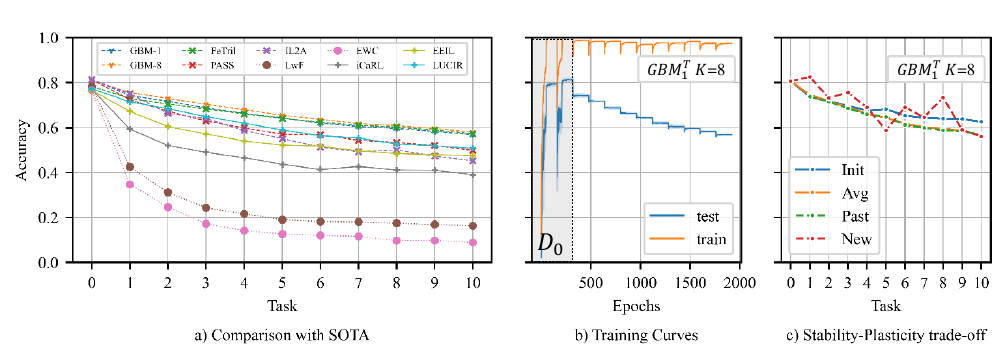}
\caption{Incremental performance of $GBM^T$ on CIFAR100, $T$=10. a) Comparison against state-of-the-art. b) Training curves on test and training sets. c) Stability-plasticity trade-off, comparing seen class accuracy (Avg) on a subset of New, Past, Initial (Init) classes.}
\label{fig:training_curves}
\end{figure*}

Furthermore our 1-prototype $GBM^T_1$ exhibits the best performance among single-prototype approaches on $T$=10 and $T$=20. For $T$=5, $GBM^T_1$ remained at $-0.9\%$ of SOTA FeTril on TinyImageNet while having features precision on 1-bit compared 32-bits floating-point. 
Finally, Figure~\ref{fig:training_curves}a) shows that the 8-prototype $GBM^T_1$ outperforms other methods on every task.

\begin{table}[htbp]
\centering
\begin{tabular}{lcl cc cc}
\hline
$D$ & $p$ or $f$ & \multicolumn{1}{l}{method} & \multicolumn{2}{c}{Task Init (0)} & \multicolumn{2}{c}{Task Final (10)} \\ \cline{4-7} 
     & \multicolumn{2}{l}{}       & Train            & Test            & Train             & Test            \\ \hline
512  & 1 & $GBM^T$  & 100.0            & 81.1            & 76.1            & 57.9            \\
     &   & $GBM^H$  & 100.0            & 81.1            & 73.5            & 54.9                \\ \hline
1024 & 2 & $GBM^T$  & 100.0            & 81.3            & 75.1            & 57.7            \\
     &   & $GBM^H$  & 100.0            & 81.0            & 73.6            & 54.6                \\ \hline
2048 & 4 & $GBM^T$  & 100.0            & 81.5            & 74.5            & 56.7            \\
     &   & $GBM^H$  & 100.0            & 80.7            & 73.6            & 54.3                \\ \hline
\end{tabular}
\caption{Investigation on the binarization method applied to Resnet-18. Results are on CIFAR100, $T$=10, $K$=8 and for both Heaviside ($GBM^H$) and Thermometer ($GBM^T$) binarizations.}
\label{tab:investigation_binarization}
\end{table}

\textbf{Sensitivity to quantization.} Table~\ref{tab:investigation_binarization} reports train and test accuracy on the initial task and the final incremental task of CIFAR100, $T$=10 for different size of binary embedding, with $K$=8 prototypes. 
For $GBM^T$ increasing the number of bits marginally improves test performance of the initial task by making $\mathcal{F}$ more expressive. On the contrary, surprisingly, a higher precision results in lower train and test accuracy on the final task. One reason can be that a lower-precision embedding regularizes the training and offers more transferable features to future incremental tasks, thus easing the optimization (train set) and enhancing generalisation (test set). For $GBM^H$, the initial and final performance is not affected by the embedding size, with less than $1\%$ difference in the performance. 
We report results for $GBM^H_{2}$, $GBM^T_{1}$ with $K$=1 and $K$=8 on CIFAR100 and TinyImageNet benchmark Table~\ref{tab:resnet_benchmark}. $GBM^T$ always yields better performance than $GBM^H$. Adding a projection to learn binary features is not beneficial. 
Figure~\ref{fig:training_curves}b) presents the training curves during all incremental retrainings for $GBM^T_1$, $K$=8, our best configuration. The train accuracy, computed on training batches (samples and pseudo-examplars), nearly always reaches $100\%$ accuracy. It suggests that the embedding binarizer does not prevent $\mathcal{F}$ and $\mathcal{G}$ to be correctly optimized despite quantization.  

\begin{table}[htbp]

\centering
\begin{tabular}{c c c}
\hline
$p$ & test acc. w/o STE phase & test acc. w/ STE phase \\ \hline
1        &  69.9    &  81.47        \\
2        &  75.0    &  81.24        \\
4        &  76.4    &  81.18        \\\hline
\multicolumn{2}{l}{ResNet-18~\cite{he2016deep}} & 81.54 \\\hline

\end{tabular}
\caption{Ablation study on the STE training phase during the initial training with Thermometer ($GBM^T$) on CIFAR100 ($T$=10).}
\label{tab:ablation_STE}
\end{table}

\textbf{Influence of training with STE.} Table~\ref{tab:ablation_STE} reports the initial test accuracy for $p$=1,2,4 with and without a second training phase with STE, in $GBM^T$. 

The results indicate that this second training phase is necessary so that a ResNet-18 with binarization reaches the same performance as without binarization, even for low-precision thermocodes on 1-bit ($p$=1). $\mathcal{F}$ can effectively be trained to mitigate quantization error and learn feature clipping.

\begin{table}[htbp]
\centering
\begin{tabular}{ccccc}
\hline
$\bm{\mu}$ init. & $\bm{\pi}$ update & $K$=1 & $K$=2 & $K$=4 \\ \hline
random   & fixed     & 65.7 & 66.0 & 66.9 \\
random   & trainable & 66.2 & 67.1  & 67.3 \\ \hline
centroid & fixed     & 66.1 & 66.7  & 67.0 \\
centroid & trainable & 66.0 & 67.4  & 67.5 \\ \hline
\end{tabular}
\caption{Influence of BMM hyper-parameters on the final average accuracy. Results are reported for $GBM^T_1$ on CIFAR100 ($T$=5).}
\label{tab:bmm_investigation}
\end{table}

\textbf{Influence of BMM initialization.} Table~\ref{tab:bmm_investigation} presents the results for the combinations of the EM-algorithm's $\bm{\mu}$ initialization and $\bm{\pi}$ update, for $K$=1,2,4. 

\begin{figure*}[htbp]
\centering
\includegraphics[trim=0 360 0 0,clip, scale = 0.97]{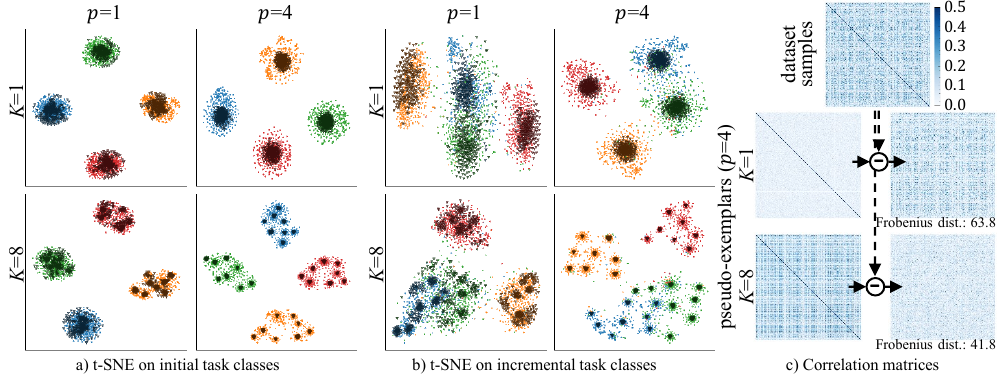}
\caption{Embedding visualization on CIFAR100 with $GBM^T_1$. a-b) t-SNE~\cite{van2008visualizing} on 4 classes, samples (light shade) and generated pseudo-exemplars (dark shade). c) Absolute difference between the correlation matrices of exemplars (dataset samples) and pseudo-exemplars.}
\label{fig:tsne_visu}
\end{figure*}

When the number of prototype augments, \textit{centroid} initialization is preferable to a random one. Even if intra-class clusters are noticeable (Figure~\ref{fig:tsne_visu}) they spread closely to the class centroid. Making $\bm{\pi}$ trainable improves final average accuracy over having them fixed, meaning that accurate mixing coefficients boost performance. For each $K$, the relative difference among all cases never exceeds $1.5\%$. Table~\ref{tab:bmm_investigation} therefore demonstrates that for such a CIL scenario, our method remains robust to a trainable $\bm{\pi}$ with a random $\bm{\mu}$ initialization. 


\textbf{Stability-Plasticity trade-off.} CIL methods should preferably ensure similar classification performance for every class, new and past. To reach this goal, the number of pseudo-exemplars generated at each iteration are so that each batch contains an equal number of elements per class. We empirically explore plasticity and stability by evaluating the accuracy on the initial, new, past, and all seen classes in Figure~\ref{fig:training_curves}.

Several observations can be made. New classes are better classified than past classes (except on task 5). Even with a fixed $\mathcal{F}$, the classifier can adapt to new classes. Nevertheless, the average relative difference between past and new accuracy is $4.5\%$, indicating a good trade-off between stability and plasticity. Finally, initial classes are better classified than past classes, hence freezing the encoder slightly prioritize ($< 6\%$) the retention of initial classes. 


\textbf{Feature Visualization.} Figure~\ref{fig:tsne_visu} illustrates the generation of pseudo-exemplars for $GBM^T$, with t-SNE~\cite{van2008visualizing} transformations. 
For a single prototype, $K$=1, on both the initial and incremental classes, pseudo-exemplars effectively approximate the sample distribution near by the centroid;
however, farther away from the centroid, pseudo-exemplars do not completely cover the sample distribution, particularly when the thermocode precision is higher, \ie $p$=4. 
This may be due to the fact that the thermometer coding may introduce feature correlations, and a BMM with a single prototype fails to represent this correlation, as illustrated in Figure~\ref{fig:tsne_visu}c). Note that the Frobenius distance between the correlation matrices of samples and pseudo-examplars remains large.
In contrast, with $K$=8, pseudo-exemplars more precisely capture sub-modalities within the class distribution, even when $p$=4. This finding is further supported by a decrease in Frobenius distance of $33\%$. Moreover, when comparing Figure~\ref{fig:tsne_visu}a) and b) we observe that the inter-class distance among incremental task classes is qualitatively smaller than that of the initial task classes. Consequently, employing multiple prototypes is advantageous, as it eases the generation of pseudo-exemplars closer to the decision boundaries.

\section{Evaluation on BNN architectures}

The literature on CIL for BNN addresses two types of networks: (1) \textbf{Hybrid BNN}, for TinyML applications, where the model size is $\sim$ $10$Mb and full-precision operations are still included~\cite{bannink2021larq}; (2) \textbf{Fully BNN}, for new hardware technologies and architecture design, where the model size is $\sim 1$Mb and binary-only arithmetic is permitted during inference. Here we show the applicability of our GBM in both distinct model types. The accuracy and memory requirements, $\mathcal{M}$, are compared only against LR approaches that scale to large datasets~\cite{Vorabbi24,basso2024class}. The memory size of the GBM prototypes depends on the number of prototypes $K$, prototype precision $q$, embedding size $D$, and total number of classes $n_c$, with $\mathcal{M}(\text{GBM})=K \times D \times n_c \times q$ bits. For binary LR, the memory size depends on the number of latent examplars $E$, with  $\mathcal{M}(\text{LR}) = E \times D \times n_c \times 1$ bits.  

\subsection{Experimental results on Hybrid BNN}\label{sec:QuickNet}

We compare to the state-of-the-art LR+CWR* method, in their network and dataset settings~\cite{Vorabbi24}. As in LR+CWR*, in GBM the QuicknetLarge backbone~\cite{bannink2021larq} is pre-trained on the ImageNet~\cite{deng2009imagenet} dataset.
$\mathcal{F}$ ends at the output of the \textit{quant\_conv2d\_30} layer, with an output dimension of $D$=$12544$ features, and the classifier, $\mathcal{G}$, consists of the next 14 layers. The benchmark is CORE50 in the \textit{NC scenario}~\cite{lomonaco2017core50}: 10 domestic objects classes learned in 9 incremental classification tasks starting with 2 objects. 

Table~\ref{tab:quicknet_benchmark} reports final accuracy for GBM and LR+CWR* for different sizes of memory. With $K$=1, GBM gives $+3.1\%$ higher final accuracy than LR+CWR* with the largest buffer, and requires $\times 4.7$ less memory space. Increasing the number of prototypes slightly increases the final accuracy by an absolute $+0.5\%$. 

\begin{table}[htbp]
\centering

\begin{tabular}{clrc}
\hline
\multicolumn{2}{c}{Method} & Memory size ($\mathcal{M}$) & Final accuracy \\ \hline
LR+CWR*     & $E$=75            &  9.4 Mb      & 69.5\%     \\
\cite{Vorabbi24}       & $E$=100           & 12.5 Mb      & 70.6\%     \\
       & $E$=150           & 18.8 Mb      & 70.8\%     \\ \hline
GBM    & $K$=1             &  4.0 Mb      & 73.9\%     \\
       & $K$=4             & 16.0 Mb      & 74.2\%     \\
       & $K$=8             & 32.0 Mb      & 74.6\%     \\ \hline
\end{tabular}
\caption{Comparison to~\cite{Vorabbi24} on CORE50 NC benchmark.}
\label{tab:quicknet_benchmark}
\end{table}

\subsection{Experimental results on Fully BNN}\label{sec:FBNN}

Prior incremental learning work is less mature on Fully BNNs. We compare GBM with LR~\cite{basso2024class} on their fully- binary VGG-like network of $\sim$$1$Mb, on the CIFAR100 dataset. 

\begin{figure}[htbp]
    \centering
    \begin{tikzpicture}
        \node[anchor=center] {\includegraphics[trim=0 7 0 0,clip, scale = 1.3]{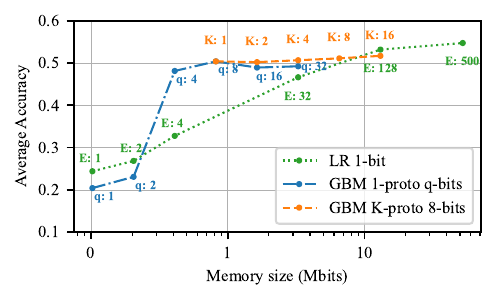}};
        
        \node[anchor=south east, font=\footnotesize] at (3.8,-0.7
        ){\cite{basso2024class}};
    \end{tikzpicture}
    \caption{Fully BNN on CIFAR-100, LR and GBM: average incremental accuracy versus required memory size.}
    \label{fig:PR_LR}
\end{figure}

An important goal is to minimize memory requirements for CIL with BNN. 
Figure~\ref{fig:PR_LR} shows the memory size required by CIL and the average incremental accuracy on CIFAR100, $T$=10. 
Three cases are compared: (1) LR with $E$ ranging from 1 to 500, (2) GBM $K$=1 with $q$ ranging from 1 to 32 to assess prototype quantization, and (3) GBM $q$=8 with $K$ ranging from 1 to 16. GBM prototypes with $q \leq 2$ cannot properly model the embeddings distribution, therefore performing worse than LR. Nevertheless, for $q \geq 3$, GBM outperforms LR with a peak at $q$=8. To achieve the same accuracy ($50.4\%$), LR requires a $\times 10$ larger replay buffer. With $q$=8 performance can still be improved up to an additional $+2\%$ by increasing the number of prototypes from 1 to 16. Prototypes quantization thus increases performance while limiting the memory required ($\leq 10$Mb). 

\section{Limitations and possible extensions}

As Fetril~\cite{petit2023fetril}, the final accuracy of our approach relies on a fixed $\mathcal{F}$, making CIL performance dependent on the initial training on $\mathcal{D}_0$.
GBM can be extended to a trainable $\mathcal{F}$, constrained with knowledge distillation \cite{zhu2021prototype,zhu2021class}. 

Furthermore, unlike standard CIL scenario benchmarks, in real-world applications, new classes might not emerge in large, discrete task datasets but as a new knowledge acquired sample-by-sample in an online manner~\cite{Hayes_2022_Embedded_CL}. Nevertheless, online versions of the EM-algorithm~\cite{cappe2009line} could extend the GBM to Online-CIL~\cite{aljundi2019gradient} and Few-Shot CIL~\cite{tao2020few}. 

On ResNet-18, we employ a linear classifier, which constrains the classes embedding to be linearly separable. Our multi-prototype approach offers the potential to approximate non-linearly separable classes distribution. Thus GBM could further benefit from more complex classifiers.

\section{Conclusion}
This paper introduces the Generative Binary Memory (GBM) method to enable pseudo-replay in Binary Neural Networks (BNNs), relying on Bernoulli Mixture Model (BMM). We demonstrate its applicability to any real-valued network, through the additional use of an embedding binarizer, achieving higher average accuracy on challenging benchmarks, such as TinyImageNet and CORE50, compared to state-of-the-art pseudo-replay and replay methods. 
Furthermore, GBM alleviates catastrophic forgetting in both large BNNs and fully binarized models, while providing superior incremental performance with a lower memory footprint, when compared to prior methods.

\section*{Acknowledgments}
This work is part of the IPCEI Microelectronics and Connectivity and was supported by the French Public Authorities within the frame of France 2030.


\bibliographystyle{unsrt}

\end{document}